\documentclass[10pt,twocolumn,letterpaper]{article}

\usepackage[pagebackref=true,breaklinks=true,letterpaper=true,colorlinks,bookmarks=false]{hyperref}
\usepackage{titling}
\usepackage[dvipsnames]{xcolor}
\usepackage{iccv}
\usepackage{times}
\usepackage{epsfig}
\usepackage{graphicx}
\usepackage{hhline}
\usepackage{amsmath}
\usepackage{amssymb}
\usepackage{comment}
\usepackage{caption}
\usepackage{stfloats}

\usepackage{algorithm}
\usepackage{algpseudocode}

\newcommand{\set}[1]{{\it{#1}}}

\newcommand\SingleLineDown[1]{%
\Statex\hspace*{-\algorithmicindent}\textbf{#1}%
\vspace*{-.7\baselineskip}\Statex\hspace*{\dimexpr-\algorithmicindent-2pt\relax}\rule{\linewidth}{0.4pt}%
}

\newcommand\DoubleLine[1]{%
\vspace*{-0.7\baselineskip}\Statex\hspace*{\dimexpr-\algorithmicindent-2pt\relax}\rule{\linewidth}{0.4pt}%
\Statex\hspace*{-\algorithmicindent}\textbf{#1}%
\vspace*{-.7\baselineskip}\Statex\hspace*{\dimexpr-\algorithmicindent-2pt\relax}\rule{\linewidth}{0.4pt}%
}

\DeclareMathOperator*{\argmax}{arg\,max}
\DeclareMathOperator*{\softmax}{softmax}
\def\v#1{\mathbf{#1}}

\iccvfinalcopy 

\def\iccvPaperID{2557} 

\ificcvfinal\fi

\begin{document}

\title{Fingerspelling recognition in the wild with iterative visual attention}

\author{Bowen Shi$^1$, Aurora Martinez Del Rio$^2$, Jonathan Keane$^2$, Diane Brentari$^2$ \\
  Greg Shakhnarovich$^1$, Karen Livescu$^1$\\
$^1$Toyota Technological Institute at Chicago, USA $^2$University of Chicago, USA\\
{\tt\small \{bshi,greg,klivescu\}@ttic.edu \ \ \ \ \ \{amartinezdelrio,jonkeane,dbrentari\}@uchicago.edu}
}

\maketitle
\begin{figure*}[!b]
  \centering
  \includegraphics[width=\linewidth]{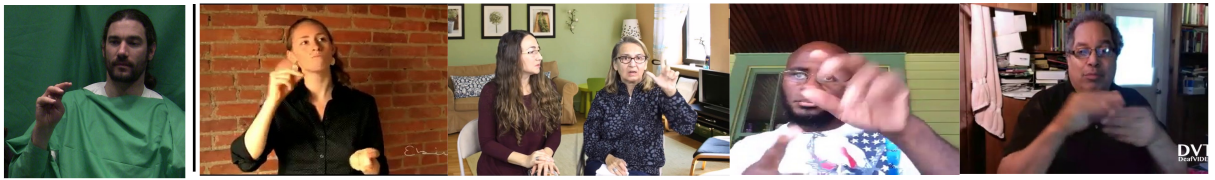}
  \caption{Fingerspelling images in studio data vs.~in the wild.  Leftmost: example frame from the ChicagoFSVid studio data set~\cite{tkim3}.  Rest:  Example frames from the
     ChicagoFSWild data set~\cite{slt_wild} (see Section~\ref{sec:data}).}\label{fig:studio_vs_wild} 
\end{figure*}

\begin{abstract}
Sign language recognition is a challenging gesture sequence recognition problem, characterized by quick and highly coarticulated motion.  
In this paper we focus on recognition of fingerspelling sequences in American Sign Language (ASL) videos collected in the wild, mainly from YouTube and Deaf social media. 
Most previous work on sign language recognition has focused on
controlled settings where the data is recorded in a studio environment
and the number of signers is limited.
Our work aims to address the
challenges of real-life data, reducing the need for detection or segmentation modules commonly used in this domain. We propose an end-to-end model
based on
an iterative
attention mechanism, without explicit hand detection or
segmentation.  Our approach dynamically focuses on increasingly high-resolution regions of interest.
It outperforms prior work by a large margin.  We also introduce a
newly collected data set of crowdsourced annotations of
fingerspelling in the wild, and 
show that performance can be further improved with this additional data set.

\end{abstract}


\section{Introduction}

Automatic recognition of sign language has the potential to overcome communication barriers for deaf
individuals.  With the increased use of online media, sign language video-based web sites (e.g., \texttt{deafvideo.tv}) are increasingly used as a platform for communication and media creation.  Sign language recognition could also enable web services like content search and retrieval in such media.

From a computer vision perspective, sign language recognition is a complex gesture recognition problem, involving quick and fine-grained motion, especially in realistic visual conditions.  It is also relatively understudied, with little existing data in natural day-to-day conditions.

\begin{figure}[t]
  \includegraphics[width=\linewidth]{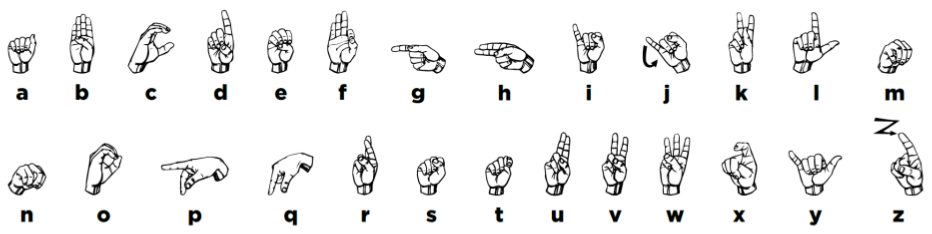}\vspace{-.3em}
  \caption{\label{fig:alphabet} The ASL fingerspelling alphabet, from \cite{jkean}.}\vspace{-1.5em}
\end{figure}

\begin{figure}[!b]
  \includegraphics[width=\linewidth]{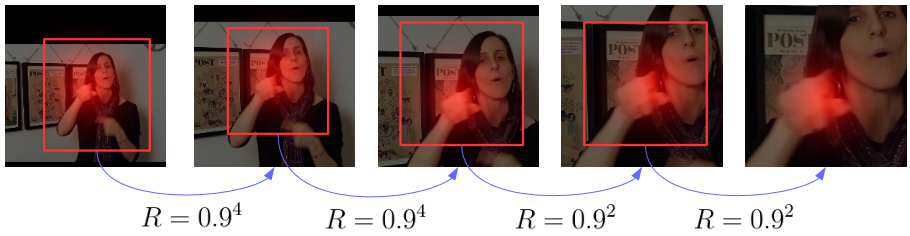}
  \caption{\label{fig:iter_zoom} Iterative attention;
  $R$=zoom factor.}
\end{figure}

\begin{figure}[!htp]
  \includegraphics[width=\linewidth]{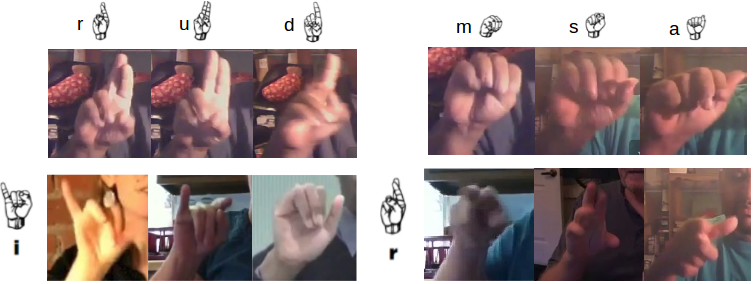}
  \caption{\label{fig:ambiguity} Ambiguity in fingerspelled handshapes. Top: different letters with similar handshapes, produced by the same signer. Bottom: same letter, different signers.}\vspace{-1em}
\end{figure}

In this paper, we study the problem of American Sign Language (ASL) fingerspelling recognition from naturally occurring sign language videos collected from web sites.
Fingerspelling is a component of ASL in which words are signed letter by letter, using an alphabet of canonical letter handshapes 
(Figure \ref{fig:alphabet}).
Words are fingerspelled mainly (but not only) when they do not have their own ASL signs, for
example technical items or proper nouns.
Fingerspelling
accounts for up to 35\%~\cite{padden} of ASL and is used frequently
for content words in social interaction or conversations involving current events or technical topics. In Deaf online media,
fingerspelling recognition is crucial as
there is often a high proportion of such content words.
Fingerspelling recognition is in some ways simpler than
general sign language recognition. In ASL, fingerspelled signs are
usually one-handed, and the hand remains in a similar position
throughout a fingerspelled sequence.  However, the task is challenging
in other ways, due to the quick, highly coarticulated, and often hard-to-distinguish finger motions, as well as motion blur in lower-quality video ``in the wild'' (Figures \ref{fig:studio_vs_wild}, \ref{fig:ambiguity}). 

Automatic sign language recognition is commonly addressed with approaches borrowed from computer vision and speech recognition. The
``front end''
usually consists of hand detection~\cite{slt_wild, jhuang} or segmentation~\cite{tkim1}, followed by visual feature extraction. Features are then passed through a sequence model, similar to ones used in speech recognition~\cite{dreuw2007speech,tkim3,deephand}.
Hand detection is
a typical first step, as sign language often involves
long sequences of large image frames. For example, in a recently introduced fingerspelling
data set~\cite{slt_wild}, a substantial proportion of sequences are more than 100 frames long, with
an average frame size of $720\times 480$, while the informative region is on
average only about $10\%$. An end-to-end recognition model on raw
image frames may have prohibitive memory requirements.

Most prior work on sign language recognition has focused on data collected in a controlled environment. Figure \ref{fig:studio_vs_wild} shows example images of fingerspelling data collected ``in the wild'' in comparison to a studio environment. Compared to studio data, naturally occurring fingerspelling images often involve more complex visual context and more motion blur, especially in the signing hand regions. Thus hand detection, an essential pre-processing step in the typical recognition pipeline, becomes more challenging.

We propose an approach for fingerspelling recognition that does not
rely on hand detection.  
Ours is an attention-based fingerspelling recognition model,
trained end-to-end from raw image frames.
We make two main contributions:  (1) We propose {\it iterative attention}, an approach for obtaining regions of interest
of high resolution with limited computation (see Figure
\ref{fig:iter_zoom}). Our model trained with iterative attention
achieves higher accuracy than the previous best
approach~\cite{slt_wild}, which requires a custom hand detector. We
further show that even
when a hand or face detector is available, our approach provides
significant added value. (2)
We introduce a new, publicly available\footnote{https://ttic.edu/livescu/ChicagoFSWild.htm} data set of crowdsourced fingerspelling video
annotations,
and show that it leads to significantly
improved fingerspelling recognition.

\section{Related Work}

Early work on sign language recognition from video\footnote{There has also been work on sign language recognition using other modalities such as depth sensors (e.g., \cite{pugeault,jhuang}).  Here we consider video-only input, as it is more abundant in naturally occurring online data.} mainly focused on isolated signs \cite{dicta_sign, asllvd_1}.
More recent work has focused on continuous sign language recognition and data sets~\cite{rwth,rwth2,tkim1,tkim3}.  Specifically for fingerspelling, the ChicagoFSVid data set includes 2400 fingerspelling sequences from 4 native ASL signers.  The RWTH-PHOENIX-Weather Corpus~\cite{rwth2} is a realistic data set of German Sign Language, consisting of sign language videos from 190 television weather forecasts.  However, its visual variability is still fairly controlled (e.g. uniform background) and it contains a small number of signers (9) signing in a fairly formal style appropriate for weather broadcasts.  In this paper we consider a constrained task (fingerspelling recognition), but with looser visual and stylistic constraints than in most previous work.  The recently introduced Chicago Fingerspelling in the Wild (ChicagoFSWild) data set \cite{slt_wild} consists of 7304 fingerspelling sequences from online videos.  This data set includes a large number of signers (168) and a wide variety of challenging visual conditions, and we use it as one of our test beds.

Automatic sign language recognition approaches often combine ideas from computer vision and speech recognition.  
A variety of sign language-specific visual pre-processing and features have been proposed in prior work, including ones based on estimated position and movement of body parts (e.g.~hand, head) combined with appearance descriptors \cite{sl_feature_1, sl_feature_2}.  Recent work has had success with convolutional neural network (CNN)-based features \cite{deephand, resign, hybrid, koller2018deep, rcn, bshi, slt_wild}.  
Much previous work on sign language recognition, and the vast majority of previous work on fingerspelling recognition, uses some form of hand detection or segmentation to localize the region(s) of interest as an initial step.
Kim {\it et al.} \cite{tkim1, tkim2, tkim3} estimate a signer-dependent skin color model 
using manually annotated hand regions for fingerspelling recognition.
Huang {\it et al.} \cite{jhuang} learn a hand detector based on Faster R-CNN~\cite{faster_rcnn} using manually annotated signing hand bounding boxes, and apply it to general sign language recognition.
Some sign language recognition approaches use no hand or pose pre-processing as a separate step (e.g.,~\cite{koller2018deep}), and indeed many signs involve large motions that do not require fine-grained gesture understanding.  However, for fingerspelling recognition it is particularly important to understand fine-grained distinctions in handshape.  Shi {\it et al.}~\cite{bshi} find that a custom-trained {\it signing} hand detector for fingerspelling recognition, which avoids detecting the non-signing hand during fingerspelling, vastly improves performance over a model based on the whole image.  
 This distinction motivates our work on iterative visual attention for zooming in on the relevant regions, without requiring a dedicated hand detector.

Once visual features are extracted, they are typically fed into sequence models such as hidden Markov models \cite{tkim1, deephand, resign}, segmental conditional random fields \cite{tkim2, tkim3}, or recurrent neural networks (RNNs) \cite{bshi, rcn, jhuang}.
In this paper, we focus on sequential models combining convolutional and recurrent neural layers due to their simplicity and recent success for fingerspelling recognition~\cite{bshi,slt_wild}.

There has been extensive work on articulated hand pose
estimation, and some models (e.g.,~\cite{simon2017hand}) have shown
real-life applicability. However, directly applying hand pose
estimation to real-life fingerspelling data is challenging.
Fingerspelling consists of quick, fine-grained movements, often
complicated by occlusion, often at low frame rates and resolutions. We
find available off-the shelf pose estimation methods too brittle to
work on our data (examples of typical failures included in the
supplementary material). 
An additional challenge to using estimated handshapes as the
  basis of fingerspelling recognition is the typically
  large discrepancies between canonical handshapes and actual
  articulation of hands in continuous signing in real-world settings.

Other related tasks include gesture recognition and action recognition. Gesture recognition is related to isolated sign recognition, which can be understood as classification of handshapes/trajectories from a sequence of frames. Most recent work \cite{gesture_1, gesture_2, gesture_3} on gesture recognition relies on depth images and typically also involves 
hand segmentation as a pre-processing step. On the other hand, action recognition \cite{action_1, action_2, action_3} is focused on classification of general video scenes based on visual appearance and dynamics.  While our task can be viewed as an example of recognizing a sequence of gestures or actions, sign language (especially fingerspelling) recognition involves discriminating fine-grained handshapes and trajectories from relatively small image regions, further motivating our approach for zooming in on relevant regions.  

Spatial attention has been applied in 
vision tasks including image captioning \cite{show_attend_tell} and
fine-grained image recognition \cite{multi_grain, Xiao2014TheAO,
  Zhao2016DiversifiedVA, Liu2016FullyCA}. Our use of attention to
iteratively zoom in on regions of interest is most similar to the work of
Fu~{\it et al.}~\cite{look_closer} using a similar ``zoom-in''
attention for image classification. Their model is trained directly from the full image, and iterative
localization provides small
gains; their approach is also
limited to a single image. In contrast, our model is applied to a
frame sequence, producing an ``attention tube'', and 
is iteratively trained with frame sequences of increasing resolution,
yielding sizable benefits.

The most closely related work to ours is that of Shi {\it et
  al.}~\cite{bshi}, which first addressed fingerseplling recognition
in the wild.  In contrast to this prior work, we propose an end-to-end
approach that directly transcribes a sequence of image frames into
letter sequences, without a dedicated hand detection step.  To our
knowledge this is the first attempt to address the continuous
fingerspelling recognition problem in challenging visual conditions, without relying on hand
detection.  
Our other main
contribution is the first successful, large-scale effort to crowdsource sign language
annotation, which significantly increases the amount of training data
and leads to a large improvement in accuracy.

\section{Task and model}

The fingerspelling recognition task takes as input a sequence of image frames (or patches) $\v I_1, \v I_2, ..., \v I_T$ and produces as output a sequence of letters $w = w_1, w_2, ..., w_K$, $K \le T$.  
Note that there is no alignment between the input and output, and typically $K$ is several times smaller than $T$ as there are several frames per letter.  We consider the lexicon-free setting, that is we do not assume a closed dictionary of allowed fingerspelled words, since fingerspelled sequences are often ones that do not occur in typical dictionaries.
Our approach for fingerspelling recognition includes the attention-based sequence model and the iterative attention approach, each described below.

\subsection{Attention-based recurrent neural network}\label{sec:arnn}

The attention-based recurrent neural network transcribes the input image sequence $\v I_1, \v I_2, ..., \v I_T$ into a letter sequence $w = w_1, w_2, ..., w_K$.
One option is to extract visual features with a 2D-CNN on individual frames and feed those features to a
recurrent neural network
to incorporate temporal
structure.  Alternatively, one can obtain a spatio-temporal representation
of the frame sequence by applying a 3D-CNN to the stacked frames. Both
approaches lack \emph{attention} -- a mechanism to focus on the
informative part of an image.  
In our case, most information is
conveyed by the hand, which often occupies only a small portion of
each frame.
This suggests using a spatial attention mechanism.

\begin{figure}[tbh]
  \includegraphics[width=\linewidth]{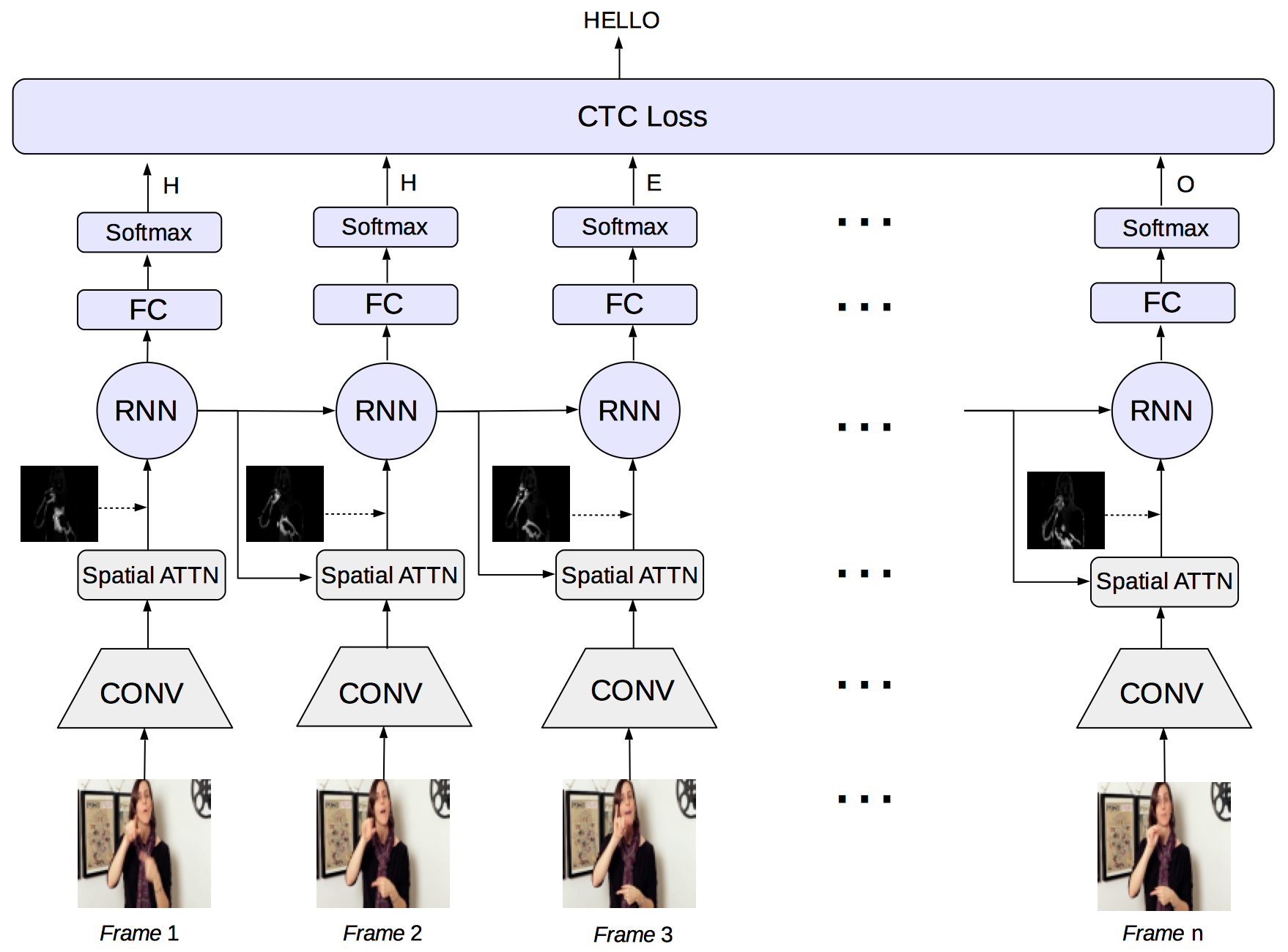}
  \caption{\label{fig:model_architecture} Recurrent CNN with attention.}\vspace{-3em}
\end{figure}
Our attention model is based on a convolutional recurrent architecture
(see Figure \ref{fig:model_architecture}). At frame $t$, a fully
convolutional neural network is applied on the image frame $\v I_t$ to
extract a feature map $\v f_t$. Suppose the hidden state of the recurrent unit at
timestep $t-1$ is $\v e_{t-1}$.  We compute the attention map $\pmb\beta_t$
based on $\v f_t$ and $\v e_{t-1}$ (where $i,j$ index spatial locations):

\begin{equation*}\label{eq:beta}
\vspace{-1em}v_{tij} = \v u_f^T \tanh\left(\v W_d \v e_{t-1}+\v W_f \v f_{tij}\right)\quad
\beta_{tij} = \frac{\exp\left(v_{tij}\right)}{\displaystyle\sum_{i, j}\exp\left(v_{tij}\right)}
\end{equation*}

The attention map $\pmb\beta_t$ reflects the
importance of features at different spatial locations to the
letter sequence. Optionally, we include a \emph{prior-based
  attention} term $\mathbf{M}$, which represents prior knowledge we
have about the importance of different spatial locations for our task.  For instance, $\mathbf{M}$
may be based on optical flow, as regions in motion are more likely than static regions to
correspond to a signing hand. The visual feature vector 
at time step $t$ is a weighted average of $\v f_{tij}$, $1\leq i\leq h,
1\leq j\leq w$, where $w$ and $h$ are the width and height of the feature
map respectively: \vspace{-.5em}
\begin{align}
  \label{eq:attn_map}\vspace{-.5em}
       \v A_t = \frac{\pmb\beta_t\odot \v M_t^\alpha}{\displaystyle\sum_{p,
        q}\beta_{tpq}M_{tpq}^\alpha},\qquad
    \v h_t=\displaystyle\sum_{i, j}\v f_{tij}A_{tij}
\end{align}
\noindent where $\mathbf{A}$ represents the (posterior) attention map and $\alpha$ controls the relative weight of the prior and attention weights learned by the model.
The state of the recurrent unit at time step $t$ is updated via
$\v e_{t} = LSTM(\v e_{t-1}, \v h^t)$
where $LSTM$ refers to a long short-term memory
unit~\cite{LSTM} (though other types of RNNs could be used here as
well).

The sequence $\mathbf{E} = (\v e_1, \v e_2, \ldots, \v e_T)$ can be viewed as high-level features for the image frames.  Once we have this sequence of features, the
next step is to decode it into a letter sequence: $(\v e_1, \v e_2,..., \v e_T)\rightarrow w = w_1, w_2, ..., w_K$. 
Our model is based on connectionist temporal classification (CTC)~\cite{graves}, which requires no frame-to-letter alignment for training.
For a sequence of visual features $\mathbf{e}$ of length $T$, we generate frame-level label posteriors via a fully-connected layer followed by a softmax, as shown in Figure \ref{fig:model_architecture}.  In CTC, the frame-level labels are drawn from $L\cup \{blank\}$, where $L$ is the true label set and $blank$ is a special label that can be interpreted as ``none of the above''.
The probability of a complete frame-level labeling $\pi = (\pi_1, \pi_2, \ldots, \pi_T)$ is then
\begin{equation}
  p(\pi|\mathbf{e}_{1:T}) = \displaystyle\prod_{t=1}^T {\softmax_{\pi_t}(\mathbf{W}^e\mathbf{e}_t+\mathbf{b}^e)}
\end{equation}
At test time, we can produce a final frame-level label sequence by taking the highest-probability label at each frame (greedy search).  Finally, the label sequence $w = w_1, w_2, ..., w_K$ is produced from the frame-level sequence $\pi$ via the CTC ``label collapsing function'' $\mathcal{B}$, which removes duplicate frame labels and then $blank$s.

At training time CTC maximizes log probability of the final label
sequence, by summing over all compatible frame-level labelings using a
forward-backward algorithm.

\textbf{Language model} In addition to this basic model, at test time
we can also incorporate a language model
providing a probability for each possible next letter given the
previous ones.
We use a beam search to find the best
letter sequence, in a similar way to decoding approaches used for
speech recognition:  The score for hypotheses in the beam is composed
of the CTC model's score (softmax output) combined with a weighted
language model probability, and an additional bias term for balancing insertions and deletions.

\subsection{Iterative visual attention via zooming in}
The signing hand(s) typically constitute only a small
portion of each frame.  In order to recognize fingerspelling sequences, the model needs to be able to reason about fine-grained motions and minor differences in handshape.  The attention mechanism enables the model 
to focus on informative regions, but high resolution is needed in order to retain sufficient information in the attended region.  One straightforward approach is to use very high-resolution input images.  However, since the
convolutional recurrent encoder covers the full image sequence, 
using large images can lead to prohibitively large memory
footprints. Using the entire frame in real-world videos also increases
vulnerability to distractors/noise.

To get the benefit of high resolution without the extra computational
burden, we propose to iteratively focus on regions within the input image frames, by
refining the attention map. Given a trained attention model
$\mathcal{H}$, we run inference with $\mathcal{H}$ on the target
image sequence $\v I_1,\v I_2,\ldots,\v I_T$ to generate the
associated sequence of posterior attention maps: $\mathbf{A}_1, \mathbf{A}_2,\ldots,\mathbf{A}_T$. We use the sequence of attention maps to obtain a new
sequence of images $\v I_1^\prime, \v I_2^\prime,\ldots,\v I_T^\prime$
consisting of smaller bounding boxes within the original images. The
fact that we extract the new frames by \emph{zooming in on the
  original images} is key, since this allows us to retain the highest
resolution available in the original video, while restricting our
attention and without paying the price for using large high
resolution frames.

We then train a new model $\mathcal{H^\prime}$ that takes $\v I_1^\prime,\v I_2^\prime,\ldots,\v I_T^\prime$ as input.  We can iterate this process, finding increasingly smaller regions of interest (ROIs).
This iterative process runs for
$S$ steps (producing $S$ trained models) until ROI images of sufficiently high resolution are obtained.  In practice, the stopping criterion for iterative attention is based on fingerspelling accuracy on held-out data.

Given a series of zooming ratios (ratios between the size of the
bounding box and the full frame) $R_1, R_2..., R_S$, the
zooming process  sequentially finds a series of bounding box sequences
$\{b_t^1\}_{1\leq t\leq T}, ..., \{b_t^S\}_{1\leq t\leq T}$. We
describe in the experiments section how we choose $R$s.

\begin{figure}[htp]
  \includegraphics[width=\linewidth]{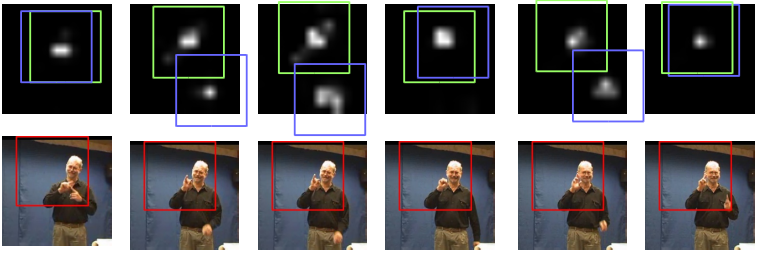}
  \caption{\label{fig:vtb} Illustration of one iteration of iterative
    attention, consisting of finding a zoomed-in ROI sequence based on
    the sequence of visual attention maps. 1st row: sequence of
    attention maps overlaid by candidate boxes of every frame. Green
    boxes are selected by dynamic programming. 2nd row: final sequence
    of bounding boxes after averaging. }
\end{figure}

This iterative process generates $S$ models.  At test time, for each input image sequence $\v I_1, \v I_2, ..., \v I_T$, the models 
$\mathcal{H}_{1:S-1}$ are run in sequence to get a sub-region sequence
$\v I^{S-1}_1, \v I^{S-1}_2, ..., \v I^{S-1}_T$.
For simplicity we just use the last model $\mathcal{H}_S$ for
word decoding based on input $\v I^{S-1}_1, \v I^{S-1}_2, ..., \v
I^{S-1}_T$
.
The iterative attention process is illustrated in
Algorithm \ref{alg:iter} and Figures~\ref{fig:iter_zoom}, \ref{fig:vtb}.

\algtext*{EndIf}%
\algtext*{EndFor}%

\begin{algorithm}[htp]
  \caption{\label{alg:iter} Iterative attention via zooming in. }
  \begin{algorithmic}[1]
    \SingleLineDown{Training, Input: $\{(\v I^{n, 0}_{1:T_n}, w^n)\}_{1\leq n\leq N}$}\;
    \For{$s \in \{1, 2,..., S\}$}
    \State Train model $\mathcal{H}_s$ with $(\v I^{n, s-1}_{1:T_n}, w^n)_{1\leq n\leq N}$\;
    \For{$n=1,...N$}
    \State Run inference on $\v I^n_{1:T_n}$ with $\mathcal{H}_s$ to
    obtain\\
    \hskip4em attention map $\v A^n_{1:T_n}$\;
    \State Solve Equation \ref{eq:vtb_max} to obtain sequence of\\
    \hskip4em bounding boxes $b^n_{1:T_n}$ \;
    \State Crop and resize $\v I^{n, 0}_{1:T_n}$ with $b^n_{1:T_n}$ to get $\v I^{n, s}_{1:T_n}$\;
    \EndFor
    \EndFor
    \State Return $\{\mathcal{H}_s\}$, $1\leq s\leq S$\;
  \end{algorithmic}
    \begin{algorithmic}[1]
    \DoubleLine{Test, Input: $\v I^0_{1:T}$, $\{\mathcal{H}_s\}$,
      $1\leq s\leq S$}
    \For{$s \in \{1, 2,..., S\}$}
    \State Run inference on $\v I^{s-1}_{1:T}$ with $\mathcal{H}_s$ to
    obtain attention\\
    \hskip4em map $\v A_{1:T}$ and predicted words $w^s$\;
    \State Solve Equation \ref{eq:vtb_max} to obtain sequence of\\
    \hskip4em bounding boxes $b_{1:T}$ \;
    \State Crop and resize $\v I^0_{1:T}$ with $b_{1:T}$ to get $\v I^{s}_{1:T}$\;
    \EndFor
    \State Return $\v w^S$
  \end{algorithmic}
  \end{algorithm}

 We next detail how bounding boxes are obtained in each
iteration of iterative attention (illustrated in Figure~\ref{fig:vtb}).  In each iteration $s$, the goal is
to find a sequence of bounding boxes $\{b_1, b_2, ..., b_T\}$ based on
the posterior attention map sequence $\{\v A_1, \v A_2, ..., \v A_T\}$ and
the zoom factor $R$, which determines the size of $b_i$ relative to the size of
$\v I^s$.  In each frame $\v I^s_t$, we put a
box of size $R_s|\v I^s|$ centered at each of the top $k$ peaks in the
attention map $\v A_t$. Each box $b_t^i$, $t\in[T]$, $i\in[k]$, is
assigned a score $a_t^i$ equal to the attention
value at its center.
We define a linking score between two bounding boxes $b_t^i$ in consecutive frames as follows:
\begin{equation}\label{eq:vtb_link}
  sc(b_t^i, b_{t+1}^j)=a_t^i+a_{t+1}^j+\lambda * IoU(b_t^i, b_{t+1}^j),
\end{equation}
\noindent where $IoU(b_t^i, b_{t+1}^j)$ is the Jaccard index (intersection over
union) of $b_t^i$ and $b_{t+1}^j$ and $\lambda$ is a hyperparameter
that trades off between the box score and
smoothness. Using $IoU$ has a smoothing effect and
ensures that the framewise bounding box does not switch between 
hands. This formulation is analogous to finding an ``action tube'' in
action recognition \cite{malik:action_tube}. Finding the sequence of
bounding boxes with highest average $s$ can be written as the optimization problem\vspace{-.5em}
\begin{equation}\label{eq:vtb_max}
  \argmax_{i_1, \ldots, i_T}\frac{1}{T}\displaystyle\sum_{t=1}^{T-1} sc(b_t^{i_t}, b_{t+1}^{i_{t+1}})\vspace{-.5em}
\end{equation}
which can be efficiently solved by dynamic programming.
Once the zooming boxes are found, we
take the average of all boxes within a sequence for further
smoothing, and finally crop the zoom-in region from the original (full
resolution) image frames to avoid any repetitive interpolation
artifacts from unnecessary resizing.  
We describe our process for
determining the zoom ratios $R_{1:S}$ in Section~\ref{sec:implementation}.

\section{Data and crowdsourced annotations}
\label{sec:data}

We use two data sets: Chicago Fingerspelling in the Wild
(ChicagoFSWild)~\cite{slt_wild}, which was carefully annotated by experts; and a crowdsourced data set we introduce here,
ChicagoFSWild+.
Both contain clips of fingerspelling sequences excised from
sign language video ``in the wild'', collected from online sources
such as YouTube and \texttt{deafvideo.tv}.  
  ChicagoFSWild contains 5455 training sequences from 87
signers, 981 development (validation) sequences from 37 signers, and 868 test sequences from 36 signers, with no overlap in signers in the three sets.

We developed a fingerspelling
video annotation interface
derived from VATIC~\cite{vatic} and have used it to collect our new data set,
ChicagoFSWild+, by crowdsourcing the annotation process via Amazon
Mechanical Turk.  Annotators are presented with one-minute ASL video clips
and are asked to mark the start and
end frames of fingerspelling within the clips (if any is present), to provide a transcription
(a sequence of English letters) for each fingerspelling sequence, but
not to align the transcribed letters to video frames.  Two annotators
are used for each clip.
The videos in ChicagoFSWild+ include varied viewpoints and styles (Figure~\ref{fig:studio_vs_wild}).

\begin{table}[htp]
  \centering
  \begin{tabular}{l|l|l|l}\hline
    {\small Gender (\%)} & {\small female \; 32.7} & {\small male \; 63.2} &
                                                                       {\small other \; 4.1} \\ \hline
    {\small Handedness (\%)} & {\small left \;\;\;\;\;\; 10.6} & {\small right
                                                   \; 86.9} & {\small other \; 2.5} \\ \hline
  \end{tabular}
  \caption{\label{tab:stats} Statistics of ChicagoFSWild+
    (train+test+dev). ``Other'' includes multiple signers, unknown, etc.  
  }\vspace{-1em}
\end{table}

ChicagoFSWild+ includes 50,402 training sequences from 216 signers (Table~\ref{tab:stats}), 3115
development sequences from 22 signers, and 1715 test sequences from 22
signers.  This data split has been
done in such a way as to approximately evenly distribute certain
attributes (such as signer gender and handedness) between the three
sets.  In addition, in order to enable clean comparisons between results
on ChicagoFSWild and ChicagoFSWild+, we used the signer labels in the two
data sets to ensure that there are no overlaps in signers between the
ChicagoFSWild training set and the ChicagoFSWild+ test set.  Finally, the annotations in the development and test sets were proofread and a single ``clean'' annotation kept for each sequence in these sets.  For the training set, no proofreading has been done and both annotations of each sequence are used.
Compared to ChicagoFSWild, the crowdsourcing setup allows us to collect
dramatically more training data in ChicagoFSWild+, with significantly
less expert/researcher effort.

\section{Experiments}
\label{sec:exp}

We report results on three evaluation sets: \set{ChicagoFSWild/dev} is used to initially assess various methods and select the
most promising ones; \set{ChicagoFSWild/test} results are directly
comparable to prior
work~\cite{slt_wild}; and finally, results on our new \set{ChicagoFSWild+/test} set
provide an additional measure of accuracy in the wild on a set more
than twice the size of \set{ChicagoFSWild/test}. 
These are the only existing data sets for sign language recognition ``in the wild'' to our knowledge.  
Performance is measured in terms of letter accuracy (in percent), computed by finding the minimum edit (Hamming) distance alignment between the hypothesized and ground-truth letter sequences.  The letter accuracy is defined as $1 - \frac{S+D+I}{N}$, where $S, D, I$ are the numbers of substitutions, insertions, and deletions in the alignments and $N$ is the number of ground-truth letters.

\subsection{Initial frame processing}
We consider the following scenarios for initial processing of the input frames:

\noindent\textbf{Whole frame} Use the full video frame, with no cropping.\\
\noindent\textbf{Face ROI} Crop a region centered on the face detection box, but 3 times larger.\\
\noindent\textbf{Face scale} Use the face detector, but instead of cropping, resize the entire frame to bring the face box to a canonical size (36 pixels).\\
\noindent\textbf{Hand ROI} Crop a region centered on the box resulting from the signing hand detector, either the same size as the bounding box or twice larger (this choice is a tuning parameter).

\subsection{Model variants}
Given any initial frame processing \textbf{X} from the list above, we compare several types of models: 

\noindent\textbf{X} Use the sequence of frames/regions in a recurrent convolutional CTC model directly (as in Figure~\ref{fig:model_architecture}, but without visual attention).  For \textbf{X = Hand ROI}, this is the approach used in~\cite{slt_wild}, the only prior work on the task of open-vocabulary fingerspelling recognition in the wild.\\
\noindent\textbf{X+attention} Use the model of Figure~\ref{fig:model_architecture}.\\
\noindent\textbf{Ours+X} Apply our iterative attention approach starting with the input produced by \textbf{X}.\\

The architecture of the recognition model described in Sec.~\ref{sec:arnn} is the same in all of
the approaches, except for the choice of visual attention model.
All input frames (cropped or
whole) are resized to a max size of 224 pixels, except for \textbf{Face scale} which yields arbitrary sized frames.
Images of higher resolution are not used due to memory constraints.

\newcommand{\nores}{{\textcolor{red}{\textbf{?}}}}

\begin{table}[t!]
  \centering{\small
  \begin{tabular}{l|c}
    \hline
    {\bf Method}& {\bf Letter accuracy (\%)}\\
    \hhline{|==|}
    Whole frame & 11.0\\
    Whole frame+attention & 23.0\\
    Ours+whole frame & 42.8\\
    Ours+whole frame\textcolor{blue}{+LM}  & 43.6\\
    \hhline{|==|}
    Face scale & 10.9\\
    Face scale+attention & 14.2\\
    Ours+face scale& 42.9\\
    Ours+face scale\textcolor{blue}{+LM}& 44.0\\
    \hhline{|==|}
    Face ROI & 27.8\\
    Face ROI+attention & 33.4\\
    Face ROI+attention\textcolor{blue}{+LM} &35.2 \\
    Ours+face ROI & 45.6\\
    Ours+face ROI\textcolor{blue}{+LM} & \textbf{46.8}\\
    \hhline{|=|=|}
    Hand ROI~\cite{slt_wild} & 41.1\\
    Hand ROI\textcolor{blue}{+LM} ~\cite{slt_wild}& 42.8\\
    Hand ROI+attention & 41.4\\
    Hand ROI+attention\textcolor{blue}{+LM} & 43.1\\
    Ours+hand ROI& 45.0\\
    Ours+hand ROI\textcolor{blue}{+LM}  & 45.9\\
    
    \hline
  \end{tabular}}
  \caption{Results on \set{ChicagoFSWild/dev}; training on
    \set{ChicagoFSWild/train}. Ours+X: iterative attention (proposed method)
    applied to input obtained with X. \textcolor{blue}{+LM}: add language model trained on
    \set{ChicagoFSWild/train}.}\vspace{-1em}
  \label{tab:dev}\vspace{-.5em}
\end{table}

\subsection{Implementation Details}\label{sec:implementation}

We use the signing \textbf{hand detector} provided by the authors
of~\cite{slt_wild}, and the two-frame motion (\textbf{optical flow}) estimation algorithm of
Farneback~\cite{Farneback}.

We use the implementation of~\cite{face_api} for \textbf{face detection},
       trained on the WIDER data set~\cite{yang2016wider}.  To save computation we run the face detector
       on one in every five frames in each sequence and interpolate to get bounding boxes for the remaining frames.       
In cases where multiple faces are detected, we form ``face tubes'' by connecting boxes in subsequent frames
with high overlap. Tubes are scored by average optical flow within an 
(expanded) box along the tube, and the highest scoring tube is
selected. Bounding box positions along the tube are averaged,
producing the final set of face detection boxes for the sequence. See
supplementary material for additional details.

\textbf{Model training} The convolutional layers of our model are based on
AlexNet~\cite{alexnet}\footnote{We do not use a deeper network like VGG~\cite{vgg}), 
  as the memory requirements are prohibitive due to its depth/stride combination when working on entire video sequences. Experiments with a relatively shallow ResNet-18 showed no improvement over the AlexNet backbone.} pre-trained on ImageNet~\cite{imagenet}. The last
max-pooling layer of AlexNet is removed so that we have a
sufficiently large feature map. When the input images are of size $224\times 224$,
the extracted feature map is of size $13\times 13$; larger inputs
yield larger feature maps.
 We include 2D-dropout layers between the last three convolutional layers with drop rate 0.2. For the RNN, we use a one-layer 
 LSTM network with 512 hidden units.  
 The model is trained with SGD, with an initial learning rate of 0.01 for 20 epochs and 0.001 for an additional 10 epochs. We use development set accuracy for early stopping. We average optical flow images at timestep $t-1$, $t$ and $t+1$, and use the magnitude as the prior map $\v M_t$ (Equation \ref{eq:attn_map}) for time step $t$. The language model is an LSTM with 256 hidden units, trained on the training set annotations.\footnote{Training on external English text is not appropriate here, since the distribution of fingerspelled words is quite different from that of English.} 
Experiments are run on an NVIDIA Tesla K40c GPU.

\textbf{Zoom ratios} For each iteration of the iterative visual
attention, we consider zoom ratios $R\in\{0.9, 0.9^2, 0.9^3, 0.9^4\}$,
and find the optimal sequence of ratios by beam search, with beam size
2, using accuracy on \set{ChicagoFSWild+/dev} as the evaluation
criterion. The parameter $\lambda$ in Equation~\eqref{eq:vtb_link} is tuned to 0.1.

\subsection{Results}
\noindent\textbf{Results on dev} Table~\ref{tab:dev} shows results on \set{ChicagoFSWild/dev} for models
trained on \set{ChicagoFSWild/train}.

First, for all types of initial frame processing, performance is improved by
introducing a standard visual attention mechanism. The whole-frame approach (whether scaled
by considering face detections, or not) is improved the
most
, since without attention too much of model capacity is wasted on
irrelevant regions; however, attention applied to whole frames remains inferior to ROI-based
methods. Using the pre-trained hand or face detector to guide the ROI
extraction produces a large boost in accuracy, confirming that
focusing the model on a high-resolution, task-relevant ROI is
important.
These ROI-based methods still benefit from adding standard
attention, but the improvements are smaller (face ROI: $5.6\%$, hand ROI: $0.3\%$). 

In contrast, our iterative attention approach, which does not
rely on any pretrained detectors, gets better performance than
detector-based methods, including the approach of~\cite{slt_wild} (\textbf{Hand ROI}), even when attention is
added to the latter (42.8\% for \textbf{Ours+whole frame} vs.~41.4\% for \textbf{Hand ROI+attention}). Our approach of
(gradually) zooming in on an ROI therefore outperforms a
signing hand detector. Specifically in \textbf{Hand ROI}, the improvement suggests signing hands can get more precisely located with our approach after initialization from a hand detector.

Finally, adding a language model yields modest accuracy improvements across the board.
The language model has a development set perplexity of 17.3, which is quite high but still much lower than the maximum possible perplexity (the number of output labels). 
Both the high perplexity and small improvement from the language model are unsurprising, since fingerspelling is often used for rare words.

\begin{table}[th!]
  \centering{\small
  \begin{tabular}{l|c|c}
    \hline
    {\bf Method}& {\bf ChicagoFSWild} & {\bf ChicagoFSWild+}\\
        \hline
    Hand ROI\textcolor{blue}{+LM}~\cite{slt_wild} & 41.9 & 41.2\\
    ~~~\textcolor{green!50!black}{+new data} & \textcolor{green!50!black}{57.5} & \textcolor{green!50!black}{58.3}\\
    \hline
    Ours+whole frame\textcolor{blue}{+LM} & 42.4& 43.8\\

    ~~~\textcolor{green!50!black}{+new data}      &\textcolor{green!50!black}{57.6} &
                                                                                               \textcolor{green!50!black}{61.0}\\
    \hline
    Ours+hand ROI\textcolor{blue}{+LM} & 42.3 & 45.9\\
    ~~~\textcolor{green!50!black}{+new data}      &\textcolor{green!50!black}{60.2} &
                                                                                               \textcolor{green!50!black}{61.1}\\

    \hline
    Ours+face ROI\textcolor{blue}{+LM} & \textbf{45.1} &
                                                         \textbf{46.7}\\
    ~~~\textcolor{green!50!black}{+new data} & \textbf{\textcolor{green!50!black}{61.2}} &\textbf{\textcolor{green!50!black}{62.3}}\\
    \hline
  \end{tabular}}
  \caption{Results on \set{ChicagoFSWild/test}
    and
    \set{ChicagoFSWild+/test}.
    Black: 
    trained
    on \set{ChicagoFSWild/train}; \textcolor{green!50!black}{Green}:
    trained on
    \set{ChicagoFSWild/train + ChicagoFSWild+/train}. \vspace{-.5em}
  }
  \label{tab:test}
\end{table}

\noindent\textbf{Results on test} We report results on
\set{ChicagoFSWild/test} for the methods that are most
competitive on dev (Table~\ref{tab:test}). All of these use some
form of attention (standard or our iterative approach) and a language
model. We again note that this table includes comparison to the only prior work
applicable to this task known to us~\cite{slt_wild}.

The combination of face-based initial ROI with our iterative attention
zooming produces the best results overall.
This is likely due to the complexity of
our image data. In cases of multiple moving objects in the same image,
the zooming-in process may fail especially in initial iterations of whole frame-based processing, when the resolution of the hand is very low because of downsampling given memory constraints.  On the other hand, the initial face-based
ROI cropping is likely to remove clutter and distractors without loss
of task-relevant information. However, even without cropping to the
face-based ROI, our approach ({\bf Ours+whole frame+LM})  still improves over
the hand detection-based one of~\cite{slt_wild}.

\noindent\textbf{Training on additional data} Finally, we report the
effect of extending the training data with the \set{ChicagoFSWild+/train}
set, 
increasing the number of training
sequences from 5,455 to 55,856.  
The crowdsourced annotations in 
ChicagoFSWild+ may be noisier, but they are much more plentiful. In
addition, 
the crowdsourced training data includes two annotations of each sequence, which can be seen as a form of natural data augmentation. 
As Table~\ref{tab:test} shows (in \textcolor{green!50!black}{green}),
all tested models benefit significantly from the new data. But the large
gap between our iterative attention approach and the hand
detector-based approach of~\cite{slt_wild} remains.
The improvement of our approach over \cite{slt_wild} applied to whole frames is \emph{larger} on the ChicagoFSWild+ test set. 
The hand detector could become less accurate due to possible domain discrepancy
between ChicagoFSWild (on which it was trained) and ChicagoFSWild+. 
In contrast, our model replacing the off-the-shelf hand detector
with an iterative attention-based ``detector'' is not influenced
by such a discrepancy.

\begin{figure}
  \includegraphics[width=\linewidth]{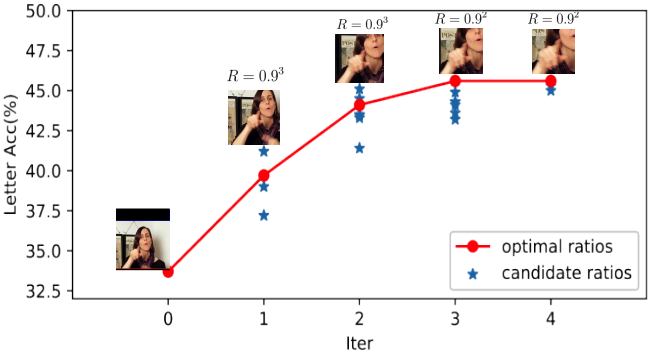}\vspace{-.5em} 
  \caption{\label{fig:acc_iter}Letter accuracy vs.~iteration in
    the {\bf Ours+face ROI} setting, showing an example ROI zooming ratio sequence found by
    beam search (shown,
    red curve).  Blue stars: accuracy with other zooming ratios considered.}\vspace{-1em}
\end{figure}

\subsection{Additional analysis}

\textbf{Effect of iterative zooming}  The results of Table \ref{tab:dev} indicate that iterative zooming gives a large performance boost over the basic model.  In both face ROI and whole frame setups,
the hand corresponds to only a small portion of the input image.  
Figure \ref{fig:acc_iter} shows how the accuracy and the input image evolve in successive zooming iterations. Though no supervision regarding the hand is used for training, the location of the signing hand is implicitly learned through the attention mechanism. Higher recognition accuracy suggests that  the learned attention locates the hand more precisely than a separately trained detector.  To test this hypothesis, we measure hand detection performance on the dev set from the hand annotation data in ChicagoFSWild.  At the same miss rate of 0.158, the average IoU's of the attention-based detector and of the separately trained hand detector are 0.413 and 0.223, respectively. Qualitatively, we also compare the sequences of signing hands output by the two detectors. 
See supplementary material for more details.

As we zoom in, two things happen:  The resolution of
the hand increases, and more of the potentially distracting background
is removed. One could achieve the former without the latter by \emph{enlarging}
the initial input by $1/R$. We compared this approach to iterative
attention, and found that (i) it was prohibitively memory-intensive (we could
not proceed past one zooming iteration), (ii) it decreased
performance, and (iii) the prior on attention became more important (see supplementary material). Therefore, iterative attention allows us to operate at much
higher resolution than would have been possible without it, and in
addition helps by removing distracting portions of the input frames.

\noindent\textbf{Robustness to face detection accuracy}  Since our best
results are obtained 
with a face
detector-based initial ROI, we investigate the sensitivity of 
the results to the accuracy of face detection, and we find that recognition performance degrades gracefully with face detector errors.
See the supplementary material for 
details and experiments.

\noindent\textbf{Timing} Our best method, {\bf Ours+face ROI+LM}, takes on average
65ms per frame.

\noindent\textbf{Human performance} We measured the letter accuracy on {\it ChicagoFSWild/test} of a native signer and two additional proficient signers.  The native signer has an accuracy of 86.1\%; the non-native signers have somewhat lower accuracies (74.3\%, 83.1\%).  These results indicate that the task is not trivial even for humans, but there is still much room for improvement from our best machine performance (61.2\%).

\section{Conclusion}
We have developed a new model for ASL fingerspelling recognition in the wild, using an
iterative attention mechanism. Our model gradually reduces its area of
attention while simultaneously increasing the resolution of its ROI
within the input frames, yielding a sequence of models of increasing
accuracy. In contrast to prior work, our approach does not rely on
any hand detection, segmentation, or pose estimation modules.
We also contribute a new data set of
fingerspelling in the wild with crowdsourced annotations, which is
larger and more diverse than any previously existing data set, and show that
training on the new data significantly improves the accuracy of all models
tested. The results of our
method on both the new data set and an existing benchmark are better
than the results of previous methods by a large margin.  We expect our iterative attention approach to be applicable to other fine-grained gesture or action sequence recognition tasks.
\vspace{-.15in}
\paragraph{Acknowledgements}
This research was supported in part by NSF grant 1433485.

{\small
  \bibliographystyle{ieee_fullname}
  \bibliography{egbib}
}

\clearpage

\appendix

\title{Supplementary Material:\\Fingerspelling recognition in the wild with iterative visual attention}
\author{Bowen Shi$^1$, Aurora Martinez Del Rio$^2$, Jonathan Keane$^2$, Diane Brentari$^2$ \\
  Greg Shakhnarovich$^1$, Karen Livescu$^1$\\
$^1$Toyota Technological Institute at Chicago, USA $^2$University of Chicago, USA\\
{\tt\small \{bshi,greg,klivescu\}@ttic.edu \ \ \ \ \ \{amartinezdelrio,jonkeane,dbrentari\}@uchicago.edu}}
\maketitle

\section{Pose estimation on fingerspelling data in the wild}
\begin{figure*}[b!]
  \includegraphics[width=\textwidth]{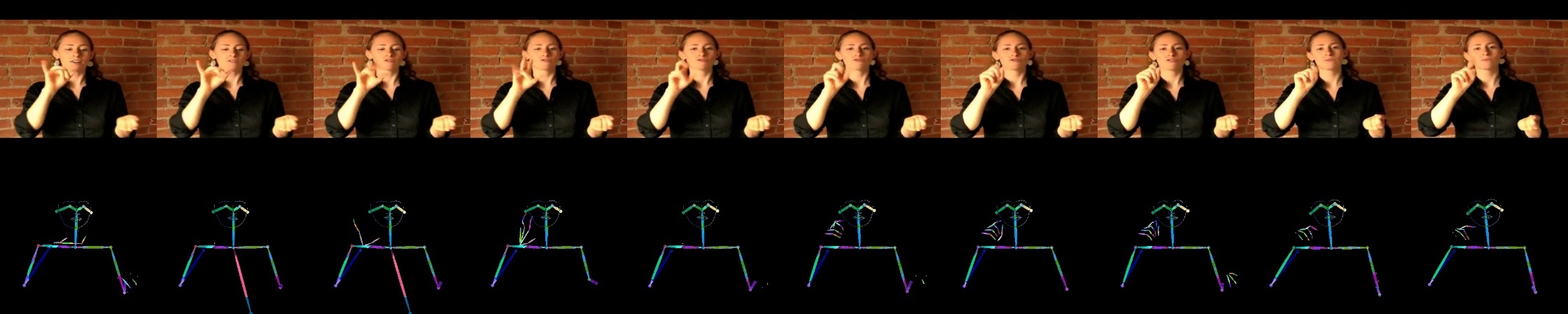} 
  \includegraphics[width=\textwidth]{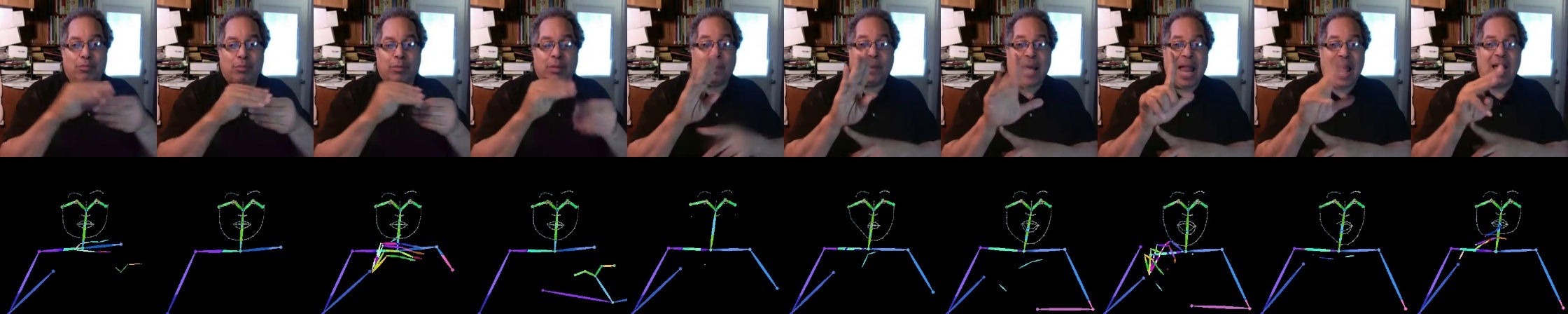} 
  \caption{\label{fig:pose}Examples of pose estimation failure on fingerspelling data from ChicagoFSWild.}
\end{figure*}

To illustrate the difficulty of using pose estimation for our
fingerspelling recognition task, we ran an off-the-shelf pose estimator, OpenPose
(\texttt{https://github.com/CMU-Perceptual- Computing-Lab/openpose}),
on our fingerspelling data.  Example results are shown in Figure \ref{fig:pose}. OpenPose is a person keypoint detection library including a hand keypoint estimation module.  
Due to the visual challenges in our fingerspelling data, signing hands are not detected in some frames. Furthermore, the hand pose is often not correctly estimated even if the signing hand is detected successfully.

\section{Face detector}

The model and training data for the face detector we use have been
described in the main paper. Here we provide additional detail on how we
apply the face detector in the {\bf face ROI} and {\bf face scale} setups, in
particular on how the ROI is extraced and scaled.

To save
computation, the face detector is run on one in every five frames per
sequence, interpolating the detections for the remaining 80\% of the frames. If only one face is detected, we take the average of all bounding boxes for the whole
sequence. In cases where multiple faces are
detected,  we first find a smooth ``face tube'' by successively taking
the bounding box in the next frame that has the highest IoU with the face
bounding box in the current frame. For every tube, a motionness score is
defined as the average value of optical flow within a surrounding
region ($3\times$ size of bounding box). Finally the tube with the highest score is selected and again the box is averaged over the whole sequence. In cases where face detection fails, we use the mean of all face bounding boxes detected in all images of the same size in the training set. We empirically observe that the failure case where no face is detected is rare ($\sim 0.5\%$  of the training set).

In the \textbf{face ROI} setting, a large region centered on the detected face bounding box is cropped and resized to serve as input.
This is because the signing hand(s) are spatially close to the face during fingerspelling. 
Specifically we crop a region centered on the bounding box which is 3
times larger. The ROI is resized with a ratio of
$\frac{224}{max(w_{roi}, h_{roi})}$ and then padded on the short side to make a squared target image of size $224\times 224$.

 In the \textbf{face scale} setting, we only scale the original frame based on the size of the face bounding box to avoid artifacts arising from cropping. The purpose of scaling is to make the scale of hands in different input sequences roughly uniform.  As our data are from videos with a large variety of viewpoints and resolutions, the scale of the hands varies over a wide range.
For instance the proportion of the hand in an image from a webcam video can be several times larger than that in an image from a third-person view.
Specifically we pre-set a base size $b$ (36 in our experiments) for the face bounding box. Input images of original size $W_{I}\times H_I$ with a bounding box of size $w_I\times h_I$ are rescaled with a ratio of $\frac{b}{max(w_I, h_I)}$. If the image area is larger than $224\times 224$ after rescaling, we further rescale by a ratio of $\alpha$ to ensure the resulting image has at most $224\times 224$ resolution due to memory constraints. $\alpha$ is multiplied in the iterative zooming-in for that input sequence.

\section{Signing hand detector}

We adopt the signing hand detector used in~\cite{slt_wild}, made
available by the authors. Unlike a
general hand detector, the objective here is to detect the
signing hand.\footnote{A large proportion of the video frames collected
  in the wild contain more than one hand.}
The detector is based on Faster R-CNN~\cite{faster_rcnn} and takes both the RGB image frame and corresponding optical flow as input.  VGG-16~\cite{vgg} is used as the backbone architecture. Unlike a general object detector, only the first 9 layers of VGG-16 are preserved and the stride of the network is reduced to 4. This is done so as to capture more fine details, since the signing hand tends to be small relative to the frame size. To enforce sequence-level smoothness, framewise bounding boxes are linked to a ``signing tube''. The linking process takes into account the IoU between bounding boxes in consecutive frames. More details on the hand detector can be found in~\cite{slt_wild}.

Apart from the original hand detector used in~\cite{slt_wild}, we also experimented with variants including using all convolutional layers of VGG-16 and concatenating feature maps in different layers to make it multi-scale as in~\cite{hand_detect_ms1,hand_detect_ms2}. We did not observe any improvement from these variants, which may be because those more complex networks suffer from overfitting due to the limited amount of hand annotations.
In addition, we notice that the majority of errors made by the hand detector consist of confusion between signing and non-signing hands instead of between hands and background objects. Typical errors can be seen in Figure~\ref{fig:comp_detector}.  Thus it is difficult to mitigate the issue of data scarcity by simply augmenting our training data with external hand datasets from other domains.

\section{Experiments on zooming vs.~enlarging, prior vs.~no prior}

We ran the following experiment to show the benefits of distraction removal obtained by the zooming employed in iterative attention, in addition to the increase in resolution. In particular, we compare the accuracy of zooming at ratio $R$ and enlarging the input images by $\frac{1}{R}$  in the {\bf face ROI} setting. For this experiment, $R$ is set to $0.9^3$, corresponding to the zooming ratio we use in the first iteration.  Comparison on smaller ratios is not feasible due to GPU memory constraints (12GB in our case).  For both zooming and enlarging, the resolution of the signing hand is the same. As can be seen from Table \ref{tab:zoom_vs_enlarge}, zooming outperforms enlarging.  When the prior map is used, the gap between the two approaches is small. This is mainly because distracting portions can be filtered via the motion-based prior in our model. The gain of zooming becomes much larger when we do not use optical flow as a complementary prior, demonstrating the benefit of distraction removal in our approach. Additionally, the motion-based prior has a negligible effect on the accuracy of our approach in this setting. 

\begin{table}[h]
  \centering
  \begin{tabular}{l|c|c}\hline
    $R=0.9^3$ & Zooming & Enlarging \\ \hline
    with prior & \textbf{39.6} & 39.3 \\ \hline
    without prior & \textbf{39.8} & 38.1 \\ \hline
  \end{tabular}
  \caption{\label{tab:zoom_vs_enlarge}Accuracy comparison between zooming and enlarging in the {\bf face ROI} setting.}
\end{table}

\section{Experiments on robustness to face detection errors }

  A face detector is used in two experimental setups: {\bf face ROI} and {\bf face scale}. To see how robust the model is to face detection errors,
we add noise to the bounding box output by the face detector. Specifically, two types of noise were separately added: size noise and position noise.
For size noise, we perturb the actual face detection boxes by multiplying the width and height of the box by
factors each drawn from $\mathcal{N}(1,\sigma_{s}^2)$.
For position noise, we add values drawn from $\mathcal{N}(0,\sigma_{p}^2)$ to the center coordinates of the face detection boxes. Note that position noise only affects the {\bf face ROI} experiments. We vary $\sigma_s$, $\sigma_p$ and show results in Table \ref{tab:size_noise}, \ref{tab:position_noise}. Overall we find that position noise has a smaller impact on accuracy compared to size noise. The {\bf face scale} setup, where no cropping is done in pre-processing, is more robust to size noise than the {\bf face ROI} setup is. Adding size noise brings a small improvement in this setting, which provides evidence that the face detector we use is not perfect. 

\begin{table}[h]
  \centering
  \begin{tabular}{l|c|c|c}\hline
    $\sigma_s$ & IoU & Face ROI & Face Scale \\ \hline
    0.0 & 1.000 & \textbf{45.6} & 42.9 \\ \hline
    0.1 & 0.858 & 45.2 & 42.7 \\ \hline
    0.2 & 0.741 & 44.7 & 43.3 \\ \hline
    0.3 & 0.641 & 44.3 & \textbf{44.0} \\ \hline
    0.4 & 0.556 & 42.6 & 43.3 \\ \hline
  \end{tabular}
  \caption{\label{tab:size_noise} Impact of size noise on letter accuracy for {\bf face ROI} and {\bf face scale} setups. IoU is measured between the perturbed and original bounding boxes.}
\end{table}

\begin{table}[h]
  \centering
  \begin{tabular}{l|c|c}\hline
    $\sigma_p$ & IoU & Face ROI \\ \hline
    0.0 & 1.000 & \textbf{45.6} \\ \hline
    0.5 & 0.780 & 45.2 \\ \hline
    1.0 & 0.621 & 45.0 \\ \hline
    1.5 & 0.499 & 44.6 \\ \hline
    2.0 & 0.402 & 44.2 \\ \hline
  \end{tabular}
  \caption{\label{tab:position_noise} Impact of position noise on letter accuracy for the {\bf face ROI} setup. Note the {\bf face scale} is not affected by position noise. IoU is measured between the perturbed and original bounding boxes.}
\end{table}

\section{Iterative attention vs.~off-the-shelf signing hand detector}
\begin{figure*}[b!]
  \begin{minipage}{0.7\textwidth}
    \vspace{0.1in}
    (1).
    \includegraphics[width=\textwidth]{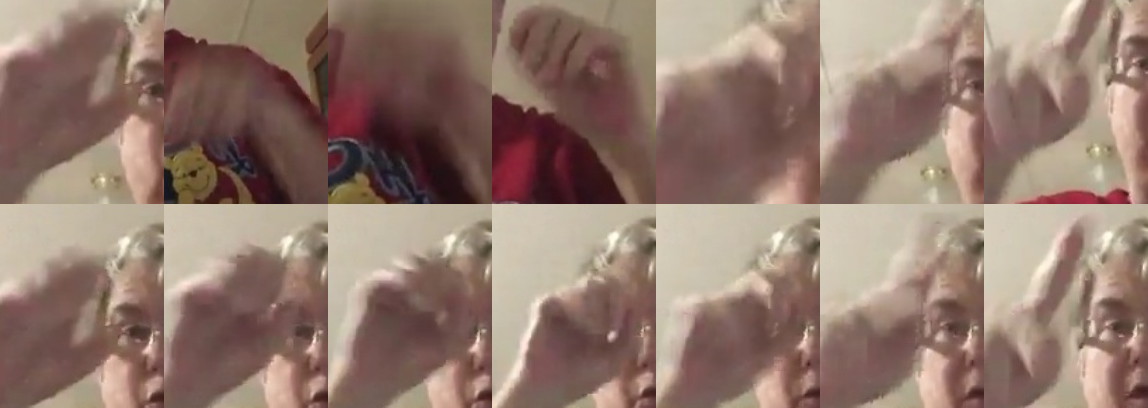}
  \end{minipage}\\
  \begin{minipage}{1.0\textwidth}
    \vspace{0.1in}
    (2).
    \includegraphics[width=\textwidth]{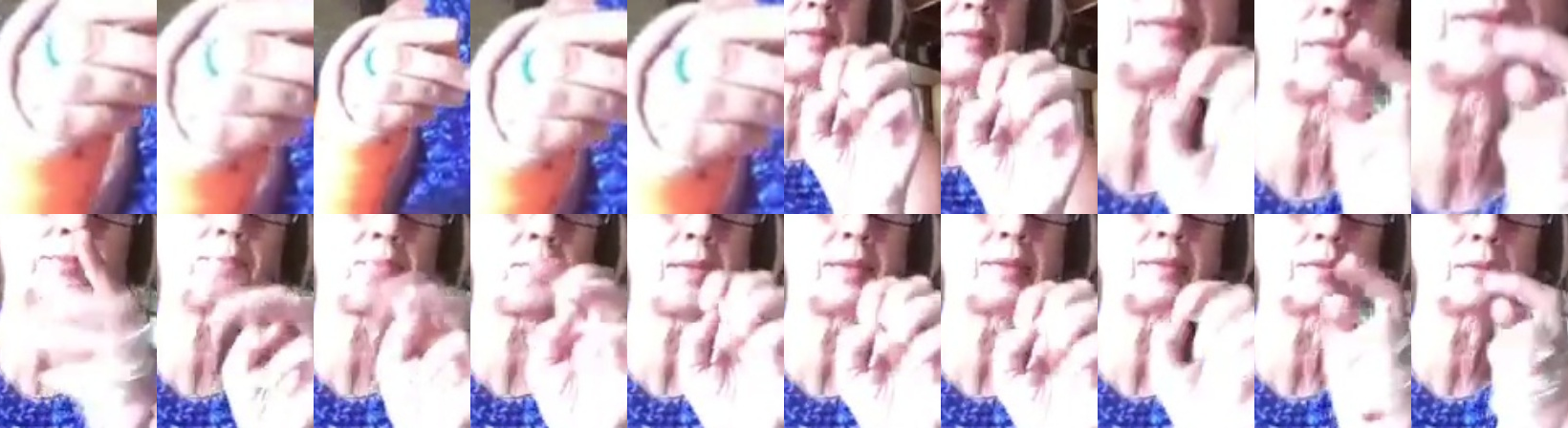}
  \end{minipage}
  \begin{minipage}{1.0\textwidth}
    \vspace{0.1in}
    (3).
    \includegraphics[width=\textwidth]{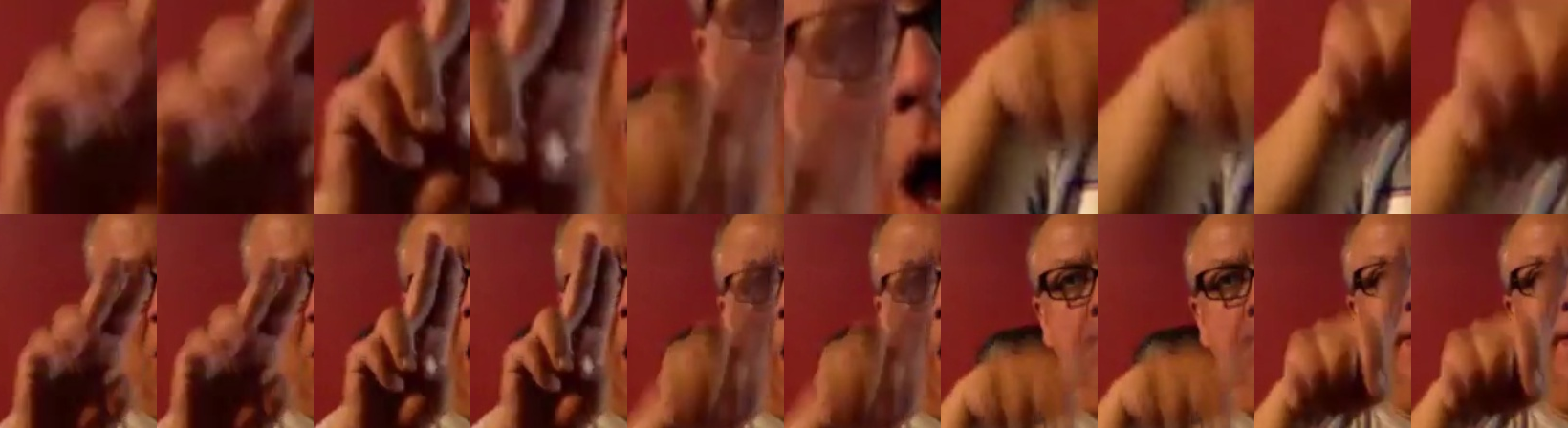}
  \end{minipage}
  \begin{minipage}{1.0\textwidth}
    \vspace{0.1in}
    (4). 
    \includegraphics[width=\textwidth]{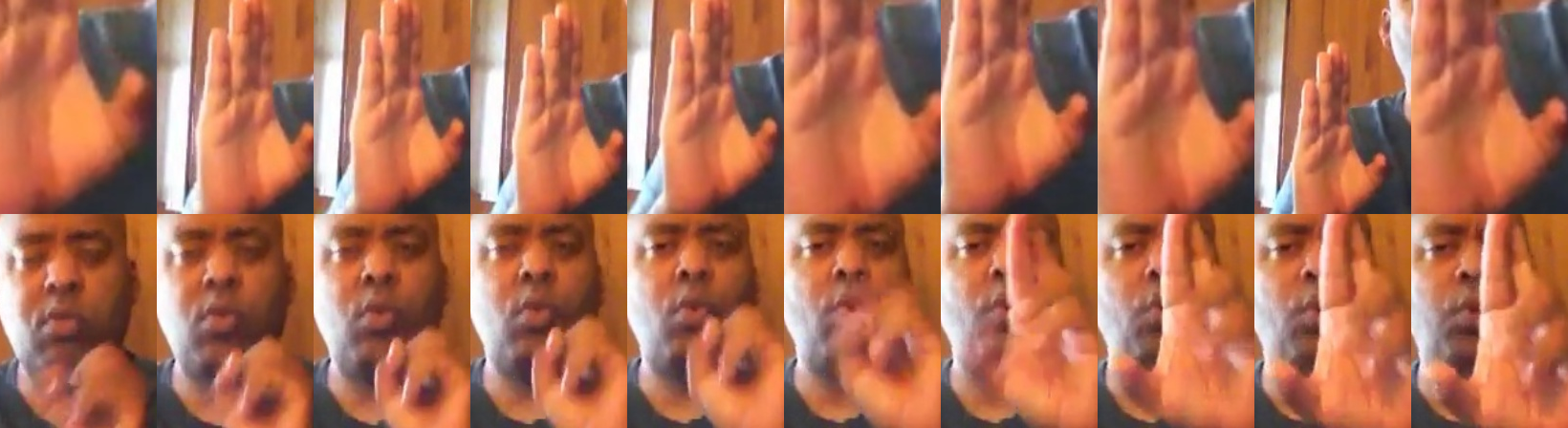}
  \end{minipage}
  \caption{\label{fig:comp_detector}Signing hands detected by the iterative attention detector vs.~the off-the-shelf signing hand detector~\cite{slt_wild}, taken from the ChicagoFSWild dev set. In each example, the upper row is from off-the-shelf detector and the lower row is from iterative attention. Signing hands are successfully detected by iterative attention in all cases. \\ Errors made by the off-the-shelf detector: In (1) and (2), bounding boxes are switched between signing and non-signing hand; in (3), the detected signing hand is incomplete; in (4), the non-signing hand is mis-detected as the signing hand. Note that sequence-level smoothing has already been incorporated in the off-the-shelf detector.}
\end{figure*}

Iterative attention serves as an implicitly learned ``detector'' of signing hands. We compare the performance of this detector
with a separately trained signing hand detector here. The signing hand detector is the one used in~\cite{slt_wild} and has been described in the previous section.
We convert the iterative attention ROI to an explicit detector through the following steps: take the input image of the last iteration, backtrack to the original image frame to get its coordinates, and use these coordinates as the bounding box. We take a model trained in the {\bf face ROI} setting and compare it with an off-the-shelf detector. Figure \ref{fig:comp_detector}  shows example sequences from the ChicagoFSWild dev set, where our approach successfully finds signing sequences while the off-the-shelf detector fails.
For quantitative evaluation, we take the dev set of hand annotation data in ChicagoFSWild, which includes 233 image frames from 19 sequences, and remove all frames with two signing hands. That amounts to 200 image frames in total. We compute average IoU and miss rate between the target bounding box and ground truth. The miss rate is defined as 1-intersection/ground-truth area. 
As the two detectors have different IoU's and miss rates, for ease of comparison we resize the bounding box of the off-the-shelf detector to keep its miss rate consistent with that of the iterative-attention detector. As is shown in Table \ref{tab:comp_detector}, our detector almost doubles the average IoU of the off-the-shelf detector at the same miss rate. Though numerical differences between IoU's may be exaggerated due to the small amount of evaluation data, the effectiveness of our approach for localization of signing hands can also be inferred from improvements in recognition accuracy. 

\begin{table}[hbt]
  \centering
  \begin{tabular}{l|c|c}\hline
     & Off-the-shelf~\cite{slt_wild} & Iterative-Attn \\ \hline
    Avg IoU & 0.213 & \textbf{0.443} \\ \hline
    Avg Miss Rate & 0.158 & 0.158 \\ \hline
  \end{tabular}
  \caption{\label{tab:comp_detector} Comparison of IoU between an off-the-shelf signing hand detector and a detector produced by iterative attention.}
\end{table}

\end{document}